\documentclass[journal]{IEEEtran}

\usepackage[utf8]{inputenc}

\usepackage{multirow}
\usepackage{booktabs,dcolumn,array}           
\usepackage{verbatim}     
\usepackage{amsmath}
\usepackage{amssymb}
\usepackage{dsfont}
\usepackage{subcaption}
\usepackage{url}
\usepackage{hyperref}

\usepackage{xcolor,colortbl}
\usepackage{pgfplots}
\usetikzlibrary{pgfplots.groupplots}

\usepackage{cases}
\usepackage{graphicx}
\usepackage{booktabs}
\usepackage{multirow}
\usepackage{scrextend}

\pgfplotsset{compat=1.14}

\begin{document}
\title{Improving Log-Cumulant Based Estimation of Roughness Information in SAR imagery }


\author{Jeova Farias Sales Rocha Neto, and Francisco Alixandre Avila Rodrigues %
\thanks{J. F. S. Rocha Neto is with the Department of Computer Science, Haverford College, Haverford, PA, 19041 USA e-mail: \url{jeovafarias@gmail.com}.}
\thanks{F. A. A. Rodrigues is with the Centro de Ciências e Tecnologia, Federal University of Cariri, Juazeiro do Norte, Brazil e-mail: \url{alixandre.avila@ufca.edu.br}.}}%

\maketitle

\begin{abstract}
Synthetic Aperture Radar (SAR) image understanding is crucial in remote sensing applications, but it is hindered by its intrinsic noise contamination, called speckle. Sophisticated statistical models, such as the $\mathcal{G}^0$ family of distributions, have been employed to SAR data and many of the current advancements in processing this imagery have been accomplished through extracting information from these models. In this paper, we propose improvements to parameter estimation in $\mathcal{G}^0$ distributions using the Method of Log-Cumulants. First, using Bayesian modeling, we construct that regularly produce reliable roughness estimates under both $\mathcal{G}^0_A$ and $\mathcal{G}^0_I$ models. Second, we make use of an approximation of the Trigamma function to compute the estimated roughness in constant time, making it considerably faster than the existing method for this task. Finally, we show how we can use this method to achieve fast and reliable SAR image understanding based on roughness information.
\end{abstract}

\begin{IEEEkeywords}
Synthetic Aperture Radar images, Bayesian Estimation, Image Analysis, Statistical Modeling.
\end{IEEEkeywords}


\section{Introduction}

\IEEEPARstart{S}{ynthetic} Aperture Radar (SAR) imaging is a powerful tool for global earth monitoring due to its advantages in operating under day or night and all-weather conditions. In has found many applications, such as change detection \cite{bouhlel2022multi}, image segmentation \cite{kayabol2015model, neto2019level} and target detection \cite{marques2011sar}, to name a few. Unfortunately, processing it is difficult due to the presence of an inherent speckle noise, since it degrades image details that are relevant for image analysis and interpretation. Therefore, statistical modeling of SAR data is a crucial step in understanding it \cite{nicolas2016statistical, frery2022sar}.  Over the years, many models were proposed to that end, with various degrees of success \cite{frery2022sar}.

In particular, the $G^0$ family of distributions \cite{frery1997model} arose to prominence in the SAR scientific community due to its flexibility in describing rough and extremely rough areas better than the $K$ distribution, besides not involving Bessel functions in their density functions. One of the most important features in these distributions is the interpretation of their $\alpha$ parameter, which varies according to the roughness of terrain being observed. Recently, directly estimating it from real data has found success in improving SAR imaging tasks \cite{rodrigues2016sar, gambini2014parameter, marques2011sar}.

In this work, we focus our attention on the method of Log-cumulants ($LCum$) to estimate $\alpha$ because of its mathematical simplicity and its previous success in preprocessing SAR imagery for segmentation \cite{rodrigues2016sar}. On the other hand, $LCum$ based estimators suffer from being computationally expensive and prone to non-convergence, especially in small sample scenarios \cite{gambini2014parameter}. Here, we aim at improving the estimation procedure in $LCum$ in two ways: (1) by making it less susceptible against estimation failures and by (2) speeding it up. For (1), we adapt the Bayesian approach pursued in \cite{cui2012bayesian} to the $G^0$ models and, for (2), we employ an approximation to the Trigamma function to expedite the inversion of a crucial equation present in $LCum$. Our final estimator is shown to, besides being more robust and faster than the traditional one, considerably reduce its estimation error.

This paper unfolds as follows. In Section II, we briefly review SAR data modeling and the method of Log-cumulants in the context of the $G^0$ distributions. We then introduce our main methodology in Section III. In Section IV, we present experimental results and briefly discuss them, and, in Section VI, we summarize our findings and conclude this work.

\section{Background} \label{sec:back}

\subsection{Statistical Models for SAR data}

SAR imagery is characterized by a typical multiplicative model in which the return in monopolarized SAR image is modeled by the product of the backscatter ($X$) and the noise ($Y$), both independent random variables. Thus, the SAR data return ($Z$) can be obtained through the product  $ Z = XY$ \cite{mejail2003classification}.

Here, we adopt the $G^{0}$ family of distributions to model the return $Z$ \cite{cui2014comparative}, a flexible, tractable and descriptive model proposed by \cite{frery1997model}, that is able to modeling homogeneous, heterogeneous and extremely heterogeneous
regions in SAR images \cite{gambini2014parameter}. In that sense, considering that the data arises from intensity evaluations at the image level, the model is called $\mathcal{G}^{0}_I$ distribution and has its probability density function (pdf) defined by:
\begin{equation}\label{gi0}
  f_{\mathcal{G}^{0}_I}(z,\theta ) = \frac{L^{L}\Gamma(L-\alpha)}{\gamma^{\alpha}\Gamma(-\alpha)\Gamma(L)} z^{L-1}(\gamma+Lz)^{\alpha - L}.
\end{equation}
For amplitude SAR data, the model is called $\mathcal{G}^{0}_A$ distribution, whose probability density function is defined as:
\begin{equation}\label{ga0}
  f_{\mathcal{G}^{0}_A}(z,\theta ) = \frac{2L^{L}\Gamma(L-\alpha)}{\gamma^{\alpha}\Gamma(-\alpha)\Gamma(L)} z^{2L-1}(\gamma+Lz^2)^{\alpha - L}
\end{equation}
For both distributions, $-\alpha,\gamma, L, z>0$ . The parameters $\alpha$ and $\gamma$ correspond to the roughness and scale, respectively, $L$ the number of looks and $\Gamma(.)$ is the gamma function. In this paper, we assume the parameter $L$ is known. 

From Eqs. (\ref{gi0}) and (\ref{ga0}), the $r$th order non-central moment for theses distributions are, respectively:
\begin{equation}\label{Eq_EGi0}
 E_{\mathcal{G}^{0}_I}[Z^r]= \left( \frac{\gamma}{L}\right)^r \frac{\Gamma(-\alpha-r)\Gamma(L+r)}{\Gamma(-\alpha)\Gamma(L)},\quad \alpha < -r
\end{equation}
and
\begin{equation}\label{Eq_EGa0}
 E_{\mathcal{G}^{0}_A}[Z^r]= \left( \frac{\gamma}{L}\right)^{\frac{r}{2}} \frac{\Gamma(-\alpha-{\frac{r}{2}})\Gamma(L+{\frac{r}{2}})}{\Gamma(-\alpha)\Gamma(L)},\quad \alpha < -{\frac{r}{2}}.
    \end{equation}


\subsection{The Log-Cumulants method (LCum) for $\mathcal{G}_I^0$ and $\mathcal{G}_A^0$ parameter estimation}

Let $Z$ be a continuous random variable with pdf $f_Z (z,\theta)$ defined over $\mathbb{R}^+$, where $\theta$ can be either a real-valued
parameter or a vector. The construction of the $LCum$ is based on the Mellin transform of $f_Z (z,\theta)$, which is defined by \cite{nicolas2002introduction}:
\begin{equation}\label{mellin}
  \phi_Z(s) \triangleq \int^\infty_0 u^{s-1}f_Z(u,\theta)\mathrm{d}u = E[Z^{s-1}].
\end{equation}
Based on the natural logarithm of $\phi_Z(s)$, the  log-moments (${m}_v$) and log-cumulants (${k}_v$) of order $v\in\mathbb{N}^*$ can be obtained by, respectively	\cite{nicolas2002introduction, REF_cheng}:
\begin{equation}
  {m}_v = \frac{\mathrm{d}^v\phi_z(s)}{\mathrm{d}s^v}\bigg|_{s=1}
\end{equation}
and 
\begin{equation}
 {k}_v \triangleq \frac{\mathrm{d}^v\psi_z(s)}{\mathrm{d}s^v}\bigg|_{s=1},
\end{equation}
\noindent with $\psi_z(s):=\log(\phi_z(s))$.

For the pdf in Eq. (\ref{gi0}) the function $\phi_Z(s)$ can be obtained using Eqs. (\ref{mellin}) and (\ref{Eq_EGi0}), which readily gives us expressions for
log-cumulants of order 1 and 2 for the $\mathcal{G}_I^0$ distribution:
\begin{equation}\label{EqSistLogCumI1}
  {k}_1 = \log \left(\frac{\gamma}{L} \right) +\Psi^0(L)-\Psi^0(-\alpha),  \\
\end{equation}
\begin{equation}\label{EqSistLogCumI2}
  {k}_2 = \Psi^{1}(L) +\Psi^{1}(-\alpha),
\end{equation}

    
\noindent where $\Psi^0(.)$ and $\Psi^1(.)$ are the digamma and trigamma functions, respectively. A similar approach can be applied to the Eq. (\ref{Eq_EGa0}) in order to calculate the first and second log-cumulants for the $\mathcal{G}_A^0$ distribution \cite{rodrigues2016sar}:
\begin{equation}\label{EqSistLogCumA1}
    2{k}_1 = \log \left(\frac{\gamma}{L} \right) +\Psi^0(L)-\Psi^0(-\alpha),\\
\end{equation}
\begin{equation}\label{EqSistLogCumA2}
    4{k}_2 = \Psi^{1}(L) +\Psi^{1}(-\alpha).
\end{equation}

Usually, systems involving Eqs. (\ref{EqSistLogCumI1})-(\ref{EqSistLogCumI2}) and (\ref{EqSistLogCumA1})-(\ref{EqSistLogCumA2}) are solved using the relation between sample log-moments and log-cumulants, which lead us to the following estimators for $k_1$ and $k_2$, respectively 
\begin{equation}\label{sampleLCum}
 \widehat{{k}}_1 = \frac{1}{n}\sum_{i=1}^n\log z_i, \quad
  \widehat{{k}}_2 = \frac{1}{n}\sum_{i=1}^n(\log z_i - \widehat{{k}}_1)^2,
\end{equation}
\noindent where $z_i$, $i\in\ \{1, 2,..., n\}$, is a sample of $Z$.

\section{Estimation Methodology} \label{sec:met}


Using Eqs. \ref{EqSistLogCumI2} and \ref{EqSistLogCumA2}, we suggest a strategy to find an approximation of the inverse of $\Psi^{1}(-\alpha)$ based on the identity:
\begin{equation}\label{eqDef}
  \Psi^{1}(- \widehat{\alpha}) = \widehat{\eta},
\end{equation}
where $\widehat{\eta} = c_{\alpha}\widehat{{k}}_2 - \Psi^{1}(L)$, where $c_{\alpha} = 1$ for the $\mathcal{G}_I^0$ and $c_{\alpha} = 4$ in $\mathcal{G}_A^0$. In order to estimate $\gamma$ as well, we can plug our estimation for $\widehat{\alpha}$ into the derivations of ${k}_1$ for the $\mathcal{G}_I^0$ and $\mathcal{G}_A^0$ models and find that:
\begin{equation}\label{eq:gamma_est}
  \widehat{\gamma} = L\exp\left(\sqrt{c_{\alpha}}{k}_1 - \Psi^0(L)+\Psi^0(-\widehat{\alpha})\right).
\end{equation}
In our work, we concentrate ourselves on the estimation of $\alpha$, leaving the analysis of the estimation of $\gamma$ to future work.

Estimating $\alpha$ using $LCum$ requires the inversion of Eq. \ref{eqDef}, which is usually done via out-of-the-shelf root finding algorithms \cite{gambini2014parameter}.  This methodology, however, as it is stated, suffers from two main problems: it requires a considerable computational processing time and fails for most cases when the estimator for $\widehat{\eta}$ returns a negative value. In other to solve such problems, we propose the following methodological improvements 

\subsection{Bayesian Approach to Reduce Failure Rate}

Inspired by the bayesian approach presented in \cite{cui2012bayesian} and the fact that $\Psi^1(x) > 0, \forall x \in \mathbb{R}^+$ \cite{abramowitz1972handbook}, we proposed  an expression for the posterior distribution of $\widehat{\eta}$ of the Eq. (\ref{eqDef}), described below.

Let $\eta_m$ be a random variable that corresponds to the posterior mean under this setting. Mathematically, the posterior PDF of $\eta_m$ given $\widehat{\eta}$ can be calculated by applying the Bayes rule in the following fashion:
\begin{equation}\label{bayes}
  f(\eta_m|\widehat{\eta}) = \frac{f(\widehat{\eta}|\eta_m)f(\eta_m)}{f(\widehat{\eta})}.
\end{equation}
Our prior knowledge on $\eta_m$ states that it should only assume positive values. Saying so, a convenient way to express that is by imposing a $\text{Uniform}(0,b)$ prior on it, i.e., $f(\eta_m; b) = 1/b \mathds{1}_{\eta_m \in (0,b)}$,
where $\mathds{1}_{\cdot}$ is the indicator function. The likelihood function  can be readily written as:
\begin{equation}
  f(\widehat{\eta}|\eta_m) = \frac{1}{\sqrt{2\pi{\sigma}^2}}\exp\left(-\frac{(\hat{\eta} - \eta_m)^2}{2{\sigma}^2}\right), 
\end{equation}
where $\sigma^2$ is the variance of $\hat{\eta}$, whose estimation is described at the end of this section. From Eq. \ref{bayes}, the posterior distribution becomes:
\begin{equation}
 f(\eta_m|\widehat{\eta}; b) = \frac{\exp\left(-\frac{(\hat{\eta} - \eta_m)^2}{2\sigma^2}\right)}{\sqrt{2\pi\sigma^2}\left(\Phi\left(\frac{\hat{\eta}}{\sigma}\right) - \Phi\left(\frac{\hat{\eta} - b}{\sigma}\right)\right)}
\end{equation}
where $\Phi(\cdot)$ is the Cumulative Distribution Function of a Standard Normal. Again following the approach in \cite{cui2012bayesian}, we compute the expected posterior mean as our estimation $\widehat{\eta_m}$ in the limit where $b \to \infty$:
\begin{align}\label{eq:esteta}
 \widehat{\eta}_m &=\nonumber \lim_{b\to \infty} \mathbb{E}[\eta_m|\widehat{\eta}; b] \\ 
 &=\nonumber \lim_{b\to \infty} \int_0^b\frac{\eta_m \exp\left(-\frac{(\hat{\eta} - \eta_m)^2}{2\sigma^2}\right)}{\sqrt{2\pi\sigma^2}\left(\Phi\left(\frac{\hat{\eta}}{\sigma}\right) - \Phi\left(\frac{\hat{\eta} - b}{\sigma}\right)\right)}\mathrm{d}\eta_m  \\
 &= \hat{\eta} + \frac{\sigma}{\sqrt{2\pi}\Phi\left(\frac{\hat{\eta}}{\sigma}\right)}\exp\left(-\frac{\hat{\eta}^2}{2\sigma^2}\right).
\end{align}

Notice that Eq. \ref{eq:esteta} proposes a correction on $\hat{\eta}$ in order to ensure that the posterior mean is positive.

In order to estimate $\sigma$, for both $\mathcal{G}_I^0$  and $\mathcal{G}_A^0$, we first notice that the randomness on $\hat{\eta}$ is uniquely due to $\widehat{k}_2$. Using Eq. \ref{sampleLCum} and defining the random variable $W =\log Z$, we can rewrite $\hat{\eta}$ as:
\begin{equation}\label{eq:expression_eta}
  \hat{\eta} = c_{\alpha}\frac{1}{n}\sum_{i=1}^n \left(w_i - \frac{1}{n}\sum_{j=1}^n w_j\right)^2  - \Psi^{1}(L).
\end{equation}

\noindent Note that the first term in the above expression is the sample variance of $W$, $S_w^2$. We can then rearrange the Eq. \ref{eq:expression_eta} as  $\hat{\eta} = c_{\alpha}S_w^2  - \Psi^{1}(L)$.
Then, the variance $\hat{\eta}$ is therefore given by the variance of $c_\alpha S_w^2$, which is computed via \cite{mood1974introduction}:
\begin{equation}
  \operatorname{Var}[\hat{\eta}] = \frac{c_\alpha^2}{n} \left(\mathbb{E}[W^4] - \frac{n-3}{n-1} \operatorname{Var}[W]^2\right).
\end{equation}
We estimate $\sigma$ in Eq. \ref{eq:esteta} as $ \sigma = \sqrt{\operatorname{Var}[\hat{\eta}]}$.

\begin{figure}[t]
\centering
 \includegraphics[width= \linewidth]{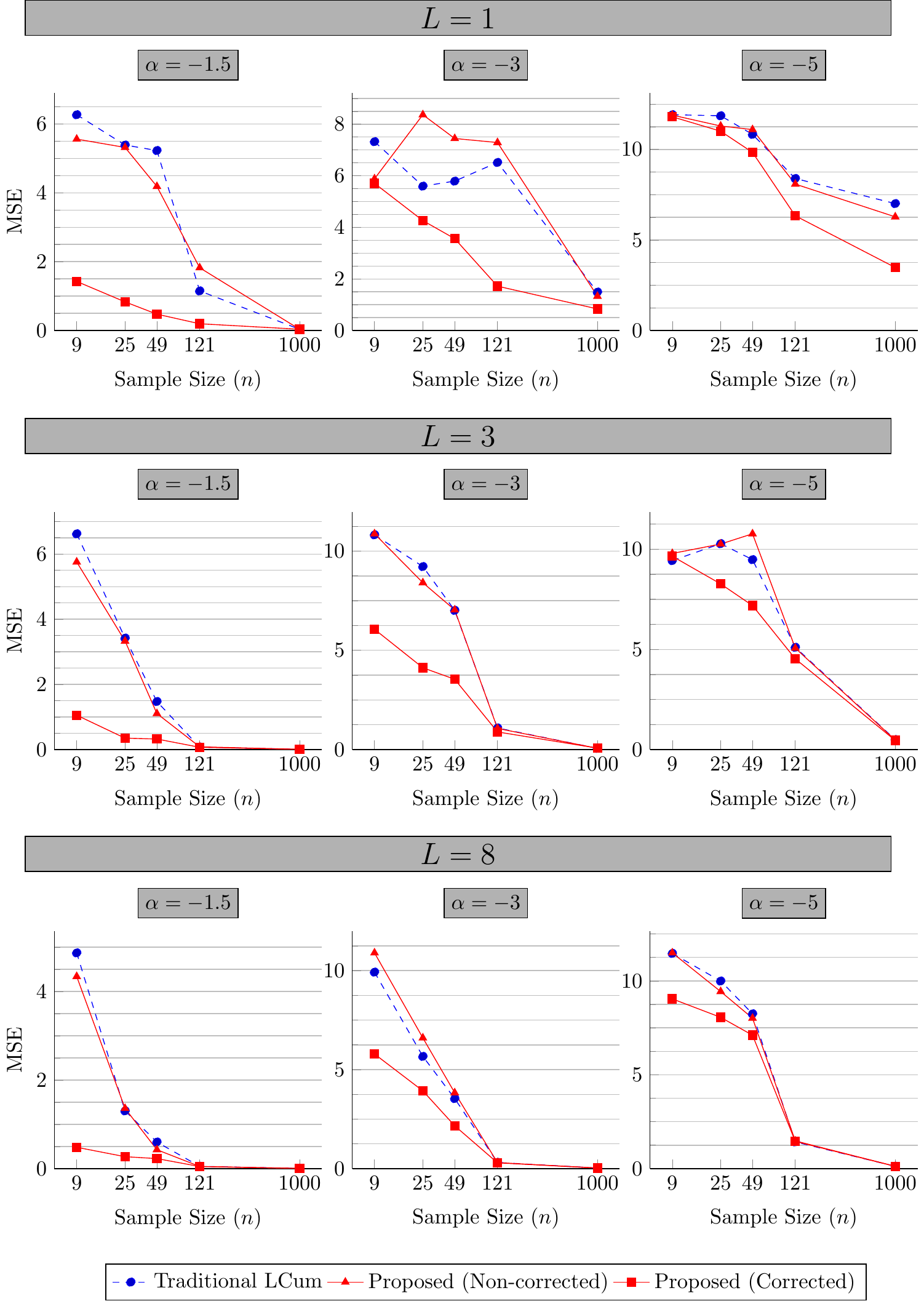}
 \caption{MSE values for the $G_I^0$ experiments.}\label{MSE_Gi0}
\end{figure}

\begin{figure}[t]
\centering
 \includegraphics[width= \linewidth]{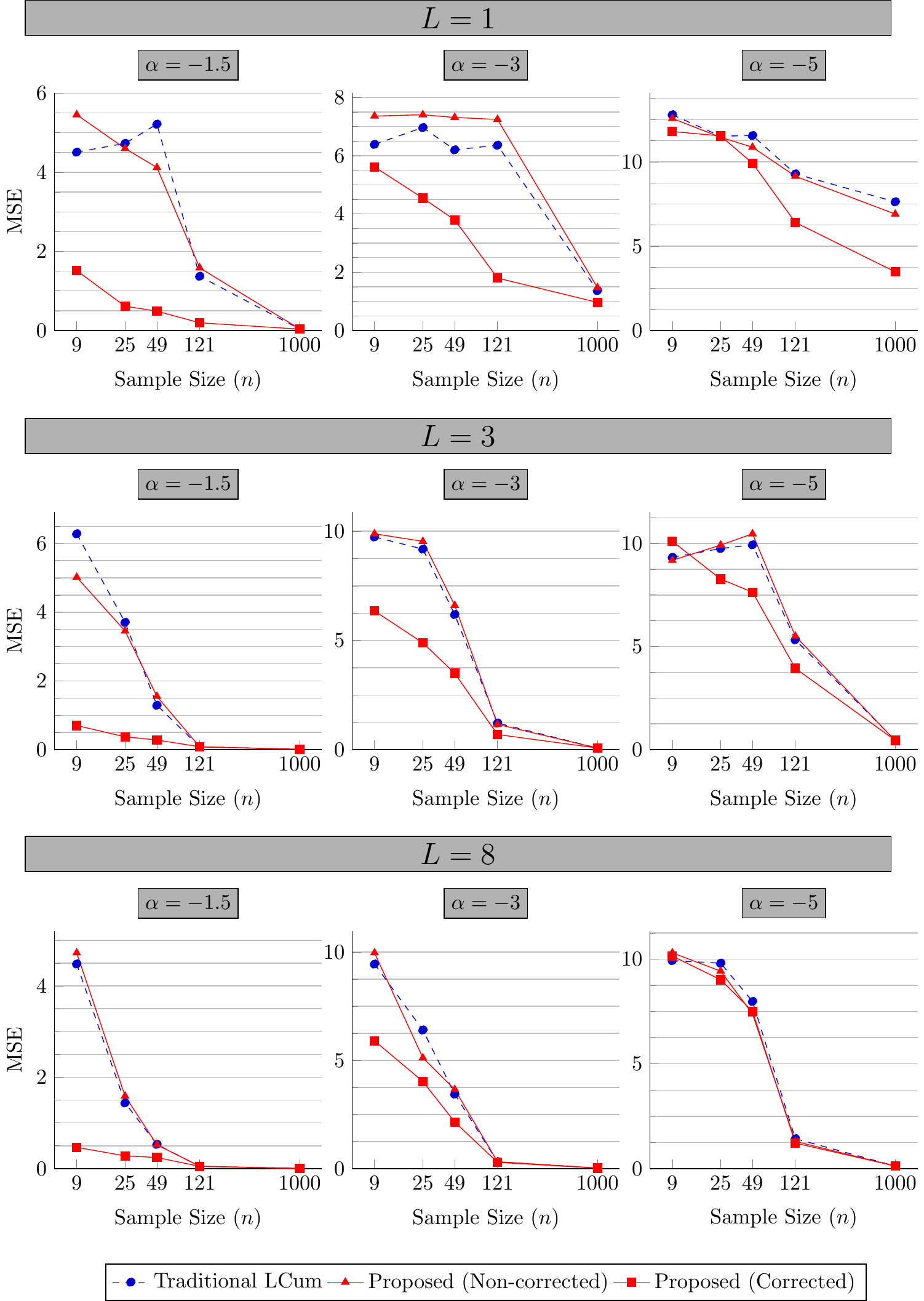}
 \caption{MSE values for the $G_A^0$ experiments.}\label{MSE_Ga0}
\end{figure}

\subsection{Trigamma Approximation to Improve Estimation Speed} \label{sec:improving_speed}
Now, back to Eq. \ref{eqDef}, we can further improve our algorithm performance by approximating the Trigamma function by an efficiently invertible one. The authors in \cite{qi2015some} propose strong bounds for $\Psi^{1}(\cdot)$ that lead us to state the following approximation on $\Psi^{1}(x)$:
\begin{align}
  \Psi^{1}(x) \approx \nonumber \frac{1}{x} +& \frac{1}{2x^2} +  \frac{1}{6x^3}-\frac{1}{30x^5} + \frac{1}{42x^7}-  \frac{1}{30x^9}.	
\end{align}

Considering that in practice $x = -\alpha \in (1, 15]$, we notice that the error for such approximation, $O\left(1/x^{11}\right)$ \cite{qi2015some} is negligible. Using our estimate for $\widehat{\eta}_{m}$ on Eq. \ref{eqDef} and the above approximation, we define the polynomial:
\begin{equation}
  P(\hat{\alpha})\triangleq 210\widehat{\eta}_{m}\hat{\alpha}^7 + 210\hat{\alpha}^6 - 105\hat{\alpha}^5 + 35\hat{\alpha}^4 - 7\hat{\alpha}^2 + 5,
\end{equation}
such that $P(\hat{\alpha}) = 0$ approximately corresponds  to  Eq. $\ref{eqDef}$. Notice that polynomial roots can be efficiently computed by the solution of an eigenvalue problem on its companion matrix.

\section{Numerical Experiments} \label{sec:Results}
\subsection{Data and Algorithmic Setup}

We demonstrate the performance of our proposed methodology using synthetic and real SAR data. The synthetic experiments were conducted on both intensity and amplitude data drawn from $G_I^0$ and $G_A^0$, respectively. To generate synthetic amplitude data, $Z_{A}$,  we used the inverse transform method according to \cite{frery2022sar}
, in which:
\begin{equation}\label{ampl}
	{Z}_{A} = \sqrt{-\frac{\gamma}{\alpha}\Upsilon^{-1}_{2L,-2\alpha}(U)},
\end{equation}
\noindent where $\Upsilon^{-1}_{2L,-2\alpha}$ is the cumulative distribution inverse function of Snedecor's $F$ law with $2L$ and $-2\alpha$ degrees of freedom, and $U$ is an uniformly distributed random variable in $(0, 1)$. And to generate intensity images, $Z_{I}$, we used the relation between amplitude and intensity data that states ${Z}_{A}^2 = {Z}_{I}$.
Without any loss of generality, for each chosen $\alpha$, we chose the scale parameter ($\gamma$) such that $E_{G_I^0}[Z_I]=1$, i.e., $\gamma = -\alpha-1$ \cite{gambini2014parameter}. 

As a real experiment, we make use of a  $600 \times 450$ pixel SAR image with $L = 4$, in L-band and HH polarization, acquired by Airborne Synthetic Aperture Radar (AIRSAR) over the San Francisco Bay \cite{rodrigues2016sar}. Our estimation methodology for this image is similar to what it is done in \cite{rodrigues2016sar}: we sweeping it with a $\ell \times \ell$ window centered on each pixel and estimate $\alpha$ for that pixel using the data from the window.

For both synthetic and real experiments, we evaluate the performance of our proposed methodology under two settings: one where we estimate $\alpha$ directly from Eq. \ref{eqDef} using our proposed approximation algorithm, as described in Section \ref{sec:improving_speed}, and another one where we first correct our $\hat{\eta}$ estimate  using Eq. \ref{eq:esteta} and then proceed with the approximation algorithm. We compare these methods to the estimation approach employed in \cite{rodrigues2016sar}, here named Traditional LCum, where the authors use MATLAB's \texttt{fsolve} function to invert the expression in Eq. \ref{eqDef}. This function, in turn, utilizes a Trust-region optimization technique \cite{conn2000trust} to solve this inversion problem. In this paper, we focus on the performance of LCum for brevity, leaving the comparisons with other estimation approaches such as Maximum Likelihood and Fractional Moments to future work. Here we focus on assessing the estimation of $\alpha$, once $\gamma$ can be simply estimated from it using Eq. \ref{eq:gamma_est}.

These experiments were run on an Intel® Core™ i5-6200U CPU \@ 2.30GHz with 8Gb of memory and the software was entirely coded in MATLAB\footnote{The code implementations from this paper can be found at \texttt{\url{github.com/jeovafarias/Improved-Roughness-Estimation-SAR}}}.

\begin{table}[t]
\caption{Failure rates (\%) on synthetic data.}\label{table_fail}
\centering
\begin{tabular}{cccccc}  
\toprule[1pt]
\multirow{2}{*}{Model}&\multirow{2}{*}{$L$}& \multicolumn{2}{c}{Proposed Method} & \multirow{2}{*}{\shortstack{Traditional \\ LCum}}\\
& & Non-corrected & Corrected &  \\\midrule
\multirow{ 3}{*}{$G_I^0$} &$1$    & 31.85  & 1.25  & 31.61 \\
&$3$    & 11.97 & 1.73 &  11.37 \\
&$8$    & 4.41 & 1.80 & 4.13 \\\midrule
\multirow{ 3}{*}{$G_A^0$} &$1$    & 30.97 & 1.40 & 32.11 \\
&$3$    & 11.62 & 2.00 & 12.22 \\
&$8$    & 4.31 & 1.27 & 4.01 \\
\bottomrule[1pt]
\end{tabular}
\end{table}

\begin{table}[t]
\caption{Average estimation time (s) on synthetic data.}\label{table_speed}
\centering
\begin{tabular}{cccccc}  
\toprule[1pt]
\multirow{2}{*}{Model}&\multirow{2}{*}{$n$} & \multicolumn{2}{c}{Proposed Method} & \multirow{2}{*}{\shortstack{Traditional \\ LCum}}\\
& & Non-corrected & Corrected &  \\\midrule
\multirow{ 3}{*}{$G_I^0$} 
& 9     & 0.56e-04 & 1.13e-04 & 2.28e-03\\
& 121   & 0.87e-04 & 1.14e-04 & 2.13e-03\\
& 1000  & 2.33e-04 & 3.43e-04 & 2.17e-03\\\midrule
\multirow{ 3}{*}{$G_I^0$} 
& 9     & 0.59e-04 & 1.05e-04 & 2.71e-03\\
& 121   & 0.91e-04 & 1.37e-04 & 2.16e-03\\
& 1000  & 2.55e-04 & 3.41e-04 & 2.18e-03\\\bottomrule[1pt]
\end{tabular}
\end{table}

\subsection{Assessment Methodology}

To analyze the robustness of the algorithms  experimented here (i.e., how likely they are to fail their estimation), we computed their failure rates according to the criteria \cite{gambini2014parameter}:
\begin{itemize}
    \item For estimates using the technique in Sec. \ref{sec:improving_speed}:
    \begin{itemize}
        \item $P(\hat{\alpha})$ presents more than one real root or none,
        \item If the real root found is lesser than -15 or positive.
    \end{itemize}
    \item For estimates arising from the Traditional LCum:
    \begin{itemize}
        \item The solver does not converge,
        \item The solution found is lesser than -15 or positive. 
    \end{itemize}
\end{itemize}

The quantitative assessment of our methods' estimation performance compared the estimated $\hat{\alpha}_i$ with their underlying values $\alpha_i$ for a sample set $i$ via Mean Square Error (MSE):
\begin{equation}\label{eq:mse}
  \operatorname{MSE} \triangleq \frac{1}{N} \sum_{i = 1}^{N} (\hat{\alpha}_i - \alpha_i)^2
\end{equation}
\noindent where $N$ is the number of estimates compared, since we only consider those that did not fail, according to our failure criteria.

Finally, we also consider the average runtime of our proposed methods for speed assessment.

\begin{figure*}
\captionsetup{justification=centering}
\begin{subfigure}{.225\textwidth}
\vspace*{-8pt}
    \includegraphics[width=\textwidth]{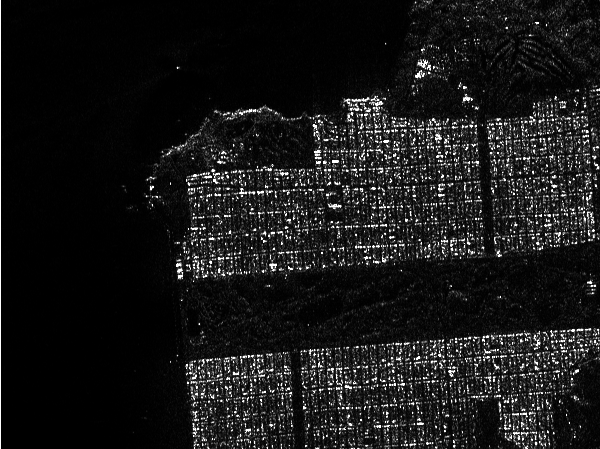}
    \caption{Original}
\end{subfigure}%
\hfill
\begin{subfigure}{.225\textwidth}
    \includegraphics[width=\textwidth]{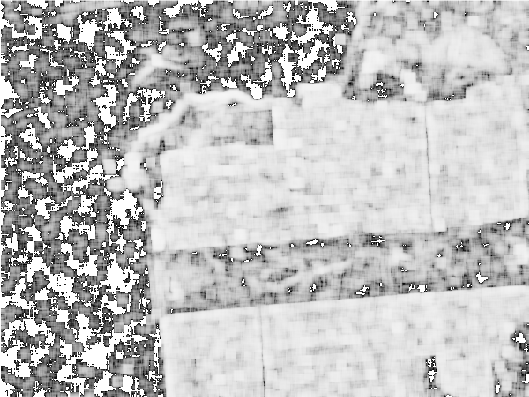}
    \caption{Traditional LCum
    \\ (ET: $604.8s$, \#Fail: 30242)}
\end{subfigure}%
\hfill
\begin{subfigure}{.225\textwidth}
    \includegraphics[width=\textwidth]{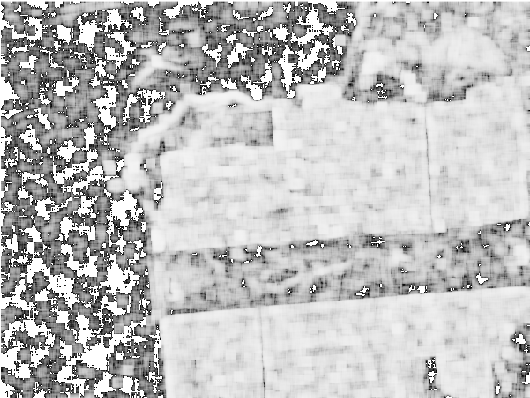}
    \caption{Proposed (non-corrected) \\ (ET: $11.2s$, \#Fail: 30304)}
\end{subfigure}
\hfill
\begin{subfigure}{.27\textwidth}
    \includegraphics[width=\textwidth]{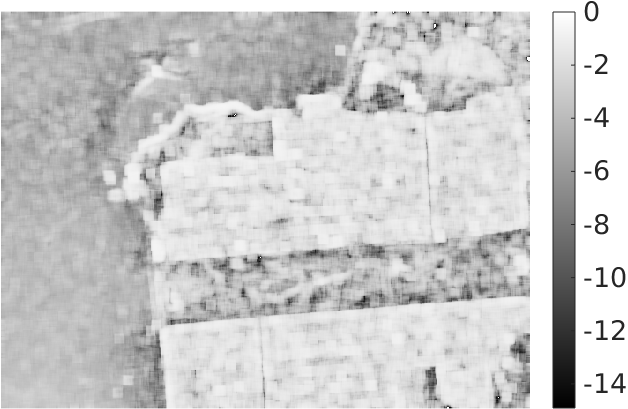}
    \caption{Proposed (corrected) \\ (ET: $39.18.3s$, \#Fail: 35)}
\end{subfigure}
        
\caption{Qualitative results for $\alpha$ estimation in a real SAR image using an $11\times 11$ window. The color bar on the right is shared by all roughness maps. ET stands for each algorithm's elapsed runtime and \#Fail is the absolute number of failures encountered when computing each map.}
\label{fig:fig_sar}
\end{figure*}

\subsection{Results on Synthetic Data} 

We start by evaluating our estimation methods on synthetic samples of size $n \in \{9, 25, 49, 121, 1000\}$,  generated using roughness $\alpha \in \{-1.5, -3.0, -5.0\}$ and number of looks $L \in \{1, 3, 8\}$. For each setting, we performed a Monte Carlo experiment with 1000 intensity and amplitude data vectors. We depict their average MSE in Figures \ref{MSE_Gi0} and \ref{MSE_Ga0}. We note that, while our proposed non-corrected estimator performs generally as well as the traditional one, our corrected estimator consistently over performs both for most scenarios.

In Table \ref{table_fail}, we use this same experimental data to compute the failure rates per value of $L$, averaging our the other parameters and averaging over experiments. Here, we again note the resemblance between our non-corrected estimator results and the ones from Traditional LCum. In both cases, we also see how increasing the number of looks helps the estimation by making the failure rate lower, a fact also observed by \cite{gambini2014parameter}. The table also shows that our corrected estimator keeps a consistent low failure rate, a clear evidence of the effectiveness of our proposed Bayesian correction approach.

Finally, we depict in Table \ref{table_speed} the average runtime of each method again using the same Monte Carlo data, but only considering the samples of sizes $n \in \{9, 121, 1000\}$. The non-corrected and corrected versions of our estimation algorithm are 6 and 40 faster than the traditional estimator. Upon inspection, most of the computation spent in our methods is due to the averaging when computing sample log-moments, which is evident when we note that the runtime increases with the sample size. That phenomenon is not explicit in the traditional LCum method, as it spends most of its time in optimizing.

\subsection{Results on Real Data} 
Figure \ref{fig:fig_sar} depicts the roughness maps generated by each method on our sample real SAR image using a $11 \times 11$ estimation window. It also presents the runtime and the absolute number of failures when computing them. For each estimation map, we set the values of $\hat{\alpha}$ to zero on locations where a failure occurred. 

Note that the estimated $\alpha$'s in all maps follow their usual understanding: higher values (greater than $-3$) are related to highly textured, urban areas; moderate values (in $[-6, -3]$) suggest forest/park zones and low values (below $-6$) correspond to textureless regions, such as the ocean \cite{gambini2014parameter}. This qualitatively  demonstrate the effectiveness of all presented methods in this experiment. Also note the similarity between the results attained by both the traditional method and our proposed non-corrected one, in terms of both estimation quality and number of failure cases. Despite this similarity, our method is around 50$\times$ faster. Furthermore, although being slower than its non-corrected counterpart, these results show that our corrected estimation encountered a negligible number of failure cases, making it more suitable to further imaging applications, such as segmentation and classification.

\section{Conclusion}
In this paper, we improve the traditional Log Cumulant-based estimation of parameters from both $G_I^0$ and $G_A^0$ in two ways. We first propose a correction to their roughness estimates using Bayesian methodology. Second, we make use of an approximation of the Trigamma function to quicken the estimation process. Our synthetic experiments demonstrate that these improvements give rise to fast and performant parameter estimation algorithms that overcome the drawbacks from its traditional approach. In fact, the approximation alone is shown to be able to speed up estimation by over 50 times, despite keeping the same estimation performance and failure rate from the traditional method. When accompanied by our Bayesian correction, the final algorithm is demonstrated to have low failure rate and higher estimation efficacy, while still being 10 times faster than its traditional counterpart. We finally also demonstrate that our proposed method is able to quickly and efficiently extract roughness maps from a real SAR image that can used for further understanding it.


\bibliography{mybib}
\bibliographystyle{IEEEtran}

\end{document}